\documentclass[review]{elsarticle}

\usepackage{amsmath,amsfonts}
\usepackage{algorithmic}
\usepackage{algorithm}
\usepackage{array}
\usepackage[caption=false,font=normalsize,labelfont=sf,textfont=sf]{subfig}
\usepackage{textcomp}
\usepackage{stfloats}
\usepackage{url}
\usepackage{verbatim}
\usepackage{graphicx}

\graphicspath{{./Images/}}

\usepackage{color}
\usepackage{multirow}
\usepackage{diagbox}
\usepackage{rotating}
\usepackage{booktabs}
\usepackage{colortbl}
\usepackage{overpic}
\usepackage[table]{xcolor}
\usepackage{textcomp}
\usepackage{contour}
\usepackage{textcomp}

\usepackage{subfig}
\usepackage{pdfpages}
\usepackage{lscape}
\usepackage{makecell}

\usepackage{setspace}
\usepackage{bm}

\definecolor{mygray}{gray}{.9}
\definecolor{myred}{RGB}{200,0,0}
\usepackage[pagebackref=false,letterpaper=true,colorlinks=true,linkcolor=myred,citecolor=blue,bookmarks=false]{hyperref}

\journal{ISPRS Journal of Photogrammetry and Remote Sensing}

\begin{document}

\begin{frontmatter}

		\title{Scribble-based Boundary-aware Network for \\Weakly Supervised Salient Object Detection in Remote Sensing Images}
		
		\author[UESTC,IIAI]{Zhou Huang}
		\ead{chowhuang@std.uestc.edu.cn}
		
		\author[IIAI]{Tian-Zhu Xiang}
		\ead{tianzhu.xiang19@gmail.com}
		
		\author[UESTC]{Huai-Xin Chen \corref{cor1}}
		\ead{huaixinchen@uestc.edu.cn}
		
		\author[MBZUAI]{Hang Dai}
		\ead{hang.dai@mbzuai.ac.ae}
		
        \address[UESTC]{School of Resources and Environment,
        University of Electronic Science \\and Technology of China, Chengdu, China}
		\address[IIAI]{Inception Institute of Artificial Intelligence (IIAI), Abu Dhabi, UAE}
		\address[MBZUAI]{Computer Vision Department, Mohamed bin Zayed University of \\Artificial Intelligence, Abu Dhabi, UAE}
		
		\cortext[cor1]{Corresponding author.}
		
		
	\begin{abstract}
		Existing CNNs-based salient object detection (SOD) heavily depends on the large-scale pixel-level annotations, which is labor-intensive, time-consuming, and expensive. By contrast, the sparse annotations (\textit{e.g.}, image-level or scribble) become appealing to the salient object detection community. However, few efforts are devoted to learning salient object detection from sparse annotations, especially in the remote sensing field. In addition, the sparse annotation usually contains scanty information, which makes it challenging to train a well-performing model, resulting in its performance largely lagging behind the fully-supervised models. 
		Although some SOD methods adopt some prior cues (\textit{e.g.}, edges) to improve the detection performance, they usually lack targeted discrimination of object boundaries and thus provide saliency maps with poor boundary localization.
		To this end, in this paper, we propose a novel weakly-supervised salient object detection framework to predict the saliency of remote sensing images from sparse scribble annotations.
		To implement it, we first construct the scribble-based remote sensing saliency dataset by relabelling an existing large-scale SOD dataset with scribbles, namely S-EOR dataset.
		After that, we present a novel scribble-based boundary-aware network (SBA-Net) for remote sensing salient object detection. 
		Specifically, we design a boundary-aware module (BAM) to explore the object boundary semantics, which is explicitly supervised by the high-confidence object boundary (pseudo) labels generated by the boundary label generation (BLG) module, forcing the model to learn features that highlight the object structure and thus boosting the boundary localization of objects. 
		Then, the boundary semantics are integrated with high-level features to guide the salient object detection under the supervision of scribble labels. 
		Extensive quantitative and qualitative evaluations on two public remote sensing SOD datasets show that the proposed method is superior to the current weakly supervised and unsupervised SOD methods and highly competitive with the existing fully supervised methods. The dataset and code will be publicly available at: \url{https://github.com/ZhouHuang23/SBA-Net}.
	\end{abstract}
		
		
			
			
			
		
	\begin{keyword}
			Salient object detection\sep saliency detection\sep scribble annotation\sep weakly supervised\sep remote sensing dataset
			
	\end{keyword}
		
\end{frontmatter}
	
	
\section{Introduction}

As a long-standing essential task in computer vision, salient object detection (SOD) aims to segment the most visually attractive regions in an image and generate a pixel-wise saliency map, which is a practical embodiment of the human visual attention mechanism~\cite{borji2019salient}. It has been widely applied to a variety of vision tasks, such as object detection \cite{zhang2019leveraging}, visual tracking \cite{zhang2017online}, person re-identification \cite{zhao2013unsupervised}, semantic segmentation \cite{hoyer2019grid}, image caption \cite{ramanishka2017top}, video compression \cite{hadizadeh2013saliency}, and camouflaged object detection \cite{fan2020camouflaged}. 
Unlike the traditional SOD based on handcrafted features, the recently deep learning based SOD models generally adopt the convolution neural networks (CNNs) or fully convolutional networks (FCNs) to learn visual saliency, which dramatically boost the detection performance. 
However, most of these methods~\cite{hou2019deeply,zhao2019egnet,pang2020multi,zhao2020suppress,zhang2020dense,huang2021semantic} strongly depend on large-scale pixel-level labeled samples for model training. The collection and manual annotation of large-scale datasets are often laborious and expensive~\cite{bearman2016s}, especially for the remote sensing images (RSIs).

\begin{figure}[bt]
	\centering
	\includegraphics[width=0.7\linewidth]{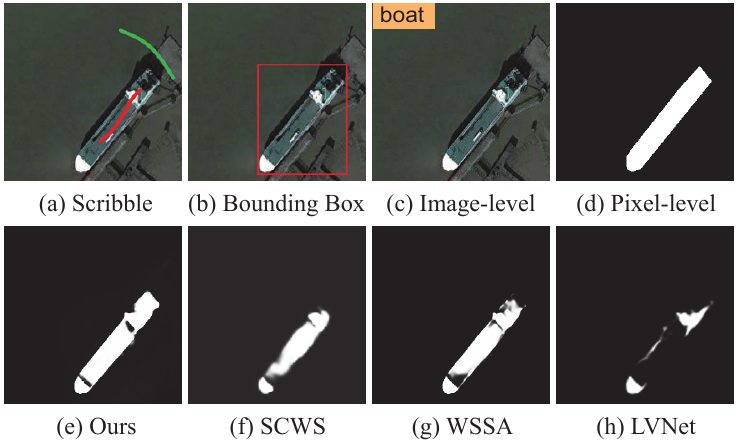}
	\caption{Visual examples of different modes of weak annotation and some salient object predictions. The first row shows different weak annotation modes, \textit{i.e.}, (a) Scribble annotations. (b) Ground-truth bounding box. (c) Ground-truth image-level annotations. (d) Ground-truth pixel-level annotations. The second row gives the saliency predictions from some weakly-supervised SOD models based on scribble annotations, \textit{i.e.}, (e) Our prediction result. (f) Result of SCWS~\cite{yu2021structure}. (g) Result of WSSA~\cite{zhang2020weakly}. (h) Result of LVNet~\cite{li2019nested}. Obviously, our method outperforms other competitors.}
	\label{fig:introduction}
\end{figure}

As an important part of salient object detection, RSI based salient object detection has been widely used in change detection~\cite{feng2018novel}, image fusion~\cite{zhang2017new}, and object detection~\cite{dong2018ship, zhang2018saliency}. 
Compared with the collection of natural images, remote sensing images are usually acquired by satellites, drones, or other aircrafts from a bird's-eye viewpoint, which leads to the great challenge for SOD from natural scene to remote sensing scene. Generally speaking, remote sensing images hold the characteristics of drastic scene changes, cluttered backgrounds, and varying scales and angles of objects~\cite{xiang2019mini, zhu2017deep}. 
It can be obviously seen that, SOD based on RSI often faces greater difficulties, such as expensive labor costs for data collection and annotation and challenging targeted model design, when compared with SOD based on natural images.

Due to the easy accessibility of sparse labels, \textit{e.g.}, only several seconds to label an image with scribbles, some attempts have been made to employ the weakly supervised annotations to train deep learning models for salient object detection. As shown in Fig.~\ref{fig:introduction}, the image-level and bounding box-level annotations are mainly used as the weak supervision for saliency detection. 
For image-level annotations~\cite{wang2017learning,li2018weakly}, owing to the complete lack of mask labels, the image-level class labels cannot provide the location and appearance information of objects, which are the very discriminative cues for object segmentation and beneficial to guide the model to distinguish salient objects from the background. Therefore, these methods based on image-level annotations are easy to be trapped in saliency maps with incomplete or low-confidence regions.
For bounding box-level annotations~\cite{liu2021weakly,zhang2020weakly}, the bounding-box labels provide position information for localizing salient objects. However, some salient objects often have irregular shapes, including winding objects and rotated objects with large aspect ratios, such as rivers, ships, and bridges. In fact, these objects are not suitable for labeling with bounding boxes, and some negative samples are easily introduced inside the bounding boxes, thus leading to interference in pixel-level object segmentation and inferior saliency prediction.
Thus, in this paper, we focus on the flexible scribble annotations as weak supervision for RSI salient object detection.

As a sparse and low-cost supervision label, the scribble annotation has received increasing attention for salient object detection and object segmentation in recent works~\cite{wang2019boundary,zhao2021weakly,valvano2021learning,hua2021semantic,wei2021scribble,yu2021structure,zhang2020weakly}.
As is known to all, the scribble annotations are too simple and sparse to convey sufficient information, \textit{e.g.}, object structure and details. Thus, it is difficult to train the deep models to focus on the entire region of the objects, which leads to the poor ability of discriminating object boundary and saliency map with blur or incomplete boundary regions.
To improve object structure in saliency predictions, some methods\cite{zhang2020dense,zhang2020weakly,zhao2019egnet,feng2019attentive,wei2021scribble} introduce the edge detection as a prior cue to supervise the model training and boost the localization of object boundary. However, these methods also increase the dependence of model training on accurate edge detection. Inferior edge detection may significantly reduce the performance of the SOD model. Besides, these methods are usually insensitive to the object boundary; that is, they are unable to select object-related critical edges as prior supervision. Thus, some noises and background edges will be easily introduced to interference model training.
Another noteworthy issue is that, to the best of our knowledge, there is not yet a sufficiently large scribble-based SOD dataset in the remote sensing community, which may hinder further research of weakly supervised learning. The scribble-based salient object detection is still an under-explored topic. 

To this end, in this paper, we propose a novel weakly-supervised salient object detection framework for remote sensing images, which learns to detect the salient objects only from sparse annotations. Firstly, due to the lack of scribble-based saliency datasets, we first re-label the existing RSI saliency dataset EORSSD~\cite{zhang2020dense} with scribbles, namely S-EOR dataset, to train our model and verify our method. Noted that the scribble annotation is conducted based on the first impression without viewing the ground-truth salient objects. Here we only use scribble annotations as ground-truth labels to train the deep model and then input remote sensing images to generate dense saliency maps during testing.
After that, to produce the high-quality saliency prediction for remote sensing images, we present a novel scribble-based boundary-aware network (SBA-Net) for remote sensing SOD. 
Specifically, we design a boundary-aware module (BAM) to distill object boundary semantics, which holds the capability to enforce the model focus on salient object structure and boost the boundary localization of objects.
Considering the differences of multi-layer features, we exploit the low-level features, which contain detailed information, to help explore boundary cues. Then we integrate the high-level features, which convey the localization semantics of salient objects, by a dense aggregation strategy (DAS) to obtain an initial saliency map. Next we combine it with the boundary cues to guide the salient object prediction under the supervision of scribble labels, which gradually restores the entire area of the salient object in a decoder network. 
The BAM is supervised explicitly by high-confidence object boundary (pseudo) labels produced from a boundary label generation (BLG) module, consisting of a classification network combined with class activation maps (CAM)~\cite{zhou2016learning}. 
As far as we know, this is the first to explore the scribble labels as supervision signals in SOD for RSI.
Overall, our main contributions are summarized as follows: 

\begin{itemize}
	\item [1)]
	We introduce the scribble annotation as weak supervision for salient object detection in remote sensing images and build the first scribble-based dataset for RSI SOD, namely S-EOR dataset.
	
	\item [2)]
	We design a novel scribble-based boundary-aware network (SBA-Net) for RSI SOD, which learns the high-quality saliency maps from remote sensing images only using the simple and sparse scribble annotations, without cumbersome pixel-wise labels.
	
	\item [3)]
	We present a boundary-aware module (BAM) to explore the boundary cues of salient objects, which can force the model to pay more attention to the object structure and then guide the high-quality saliency prediction. The BAM is supervised explicitly by the designed boundary label generation (BLG) module.  
	
	\item [4)]
	Extensive experimental results on two challenging RSI SOD datasets demonstrate that our method outperforms other weakly-supervised/unsupervised competitors, and is on par with several fully-supervised models, using six widely used evaluation metrics.
	
\end{itemize}

\section{Related Work}

\subsection{Fully Supervised Salient Object Detection}

Deep fully-supervised SOD models have been widely studied and achieved impressive  progress~\cite{borji2019salient,wang2021salient,fan2021salient}. Most SOD methods~\cite{deng2018r3net,hou2019deeply,hu2018recurrently,zhao2019pyramid,pang2020multi,zhao2020suppress} mainly focus on solving the integration of multi-level features to generate a salient map with accurate location and internal consistency. Other works are devoted to studying how to use multi-modal information, such as depth cues~\cite{zhou2021rgb,fan2020rethinking,huang2021multi} or thermal infrared cues~\cite{tu2020rgbt,zhou2021ecffnet}, as auxiliary inputs of the models to alleviate the defects of individual RGB sources, such as low-light environments and similar texture scenes. It is noted that, some methods~\cite{zhang2020dense,zhang2020weakly,zhao2019egnet,feng2019attentive} also introduce the edge detection as a joint or auxiliary task for SOD, which make the models pay more attention to the object structure and thus improve the localization of salient objects. Although these methods show impressive performance, they all rely on expensive dense pixel-level annotations for training.

\subsection{Weakly Supervised Salient Object Detection}

To alleviate the demand for accurate pixel-level annotations and reduce the cost of dataset construction, some SOD methods attempt to introduce some low-cost annotations, such as image-level or scribble labels, for saliency detection, namely weakly-supervised/unsupervised SOD.
To the best of our knowledge, Wang \textit{et al.}~\cite{wang2017learning}  first proposed a deep model for learning saliency from weakly supervised annotations, \textit{i.e.}, image-level labels. The proposed method adopts a joint training strategy composed of a foreground inference network and a fully convolutional network, which improves the foreground detection of unforeseen categories. 
Considering multiple available sources of weak supervision, Zeng \textit{et al.}~\cite{zeng2019multi} proposed a unified weak supervision framework to learn the saliency from diverse supervision sources, \textit{e.g.}, category labels, captions, and noisy pseudo labels. The method utilizes a classification network and a caption generation network to generate pixel-level pseudo labels to train a saliency prediction network for the prediction of saliency maps.
In recent works, Zhang \textit{et al.}~\cite{zhang2020weakly} is the first to introduce scribble annotations for saliency prediction and proposed a weakly-supervised salient object detection model with an auxiliary edge detection network and a gated structure-aware loss to produce high-quality saliency maps. In order to make it easier to train the scribble-supervised model, Yu \textit{et al.}~\cite{yu2021structure} proposed a local coherence loss based on image features and pixel distance to achieve the one-round end-to-end model training.

It is observed that studies on scribble-based weakly-supervised SOD still remain scarce. It is in its infancy, thus leaving considerable room for improvement. To our knowledge, in the remote sensing community, learning saliency from scribble annotations has not been challenged yet. Therefore, in this paper, we build the first scribble-based dataset for remote sensing image SOD and design a novel boundary-aware network for learning the saliency of remote sensing images from scribble annotations. Unlike \cite{yu2021structure} and \cite{zhang2020weakly}, we present a boundary-aware module to explore the boundary cues from low-level features and images to improve the boundary localization of salient objects, which is explicitly supervised by a boundary label generation module consisting of a classification network with class activation maps. The well-designed module can distinguish the critical object-boundary cues to boost the accurate salient object detection.

\subsection{Semantic Segmentation Based on Scribble Annotation}

The scribble annotation is also applied in weakly supervised semantic segmentation. For natural images, Lin \textit{et al.}~\cite{lin2016scribblesup} introduced scribble annotation as a weak supervision signal for semantic segmentation for the first time. The proposed method adopts a graph-based model to propagate supervision signals from scribble to unlabeled pixels. To mine the limited annotation cues from scribble labels, Wang \textit{et al.}~\cite{wang2019boundary} proposed a prediction refinement module and an edge regression module, among which edge regression learns the class-independent boundary mapping for branch supervision.
For remote sensing images, Wei \textit{et al.}~\cite{wei2021scribble} proposed a boundary detection branch along with the semantic segmentation network for scribble-based RSI road extraction. In \cite{hua2021semantic}, a weak supervision method based on multi-mode sparse annotation was studied for RSI semantic segmentation. 
For medical images, Dorent \textit{et al.}~\cite{dorent2020scribble} introduced domain adaptation based on scribble labels for co-segmentation. Valvano \textit{et al.}~\cite{valvano2021learning} proposed the multi-scale adversarial attention gates to learn segmentation from scribbles. 
Compared with the weakly-supervised semantic segmentation, which uses full-class scribble annotations, the scribble-based SOD only uses two labels. Except for the foreground label, the background label often covers single- or multi-class objects. Thus, to some extent, the scribble-based SOD task may be more challenging.

\subsection{Salient Object Detection for RSI}

Early studies for RSI saliency detection focused on handcrafted features. Zhang \textit{et al.}~\cite{zhang2018saliency} proposed an RSI saliency analysis model based on multispectral image clustering and panchromatic image co-occurrence histogram. 
Huang \textit{et al.}~\cite{huang2021contrast} proposed contrast-weighted atoms for saliency detection using sparse representation and dictionary learning.
Recently, some deep learning based methods have been proposed for saliency detection. Li \textit{et al.}~\cite{li2019nested} built the first public optical RSI SOD dataset and designed an end-to-end network consisting of a two-stream pyramid module and an encoder-decoder module with nested connections for saliency detection. Zhang \emph{et al.}~\cite{zhang2020dense} explored the fusion of high- and low-level attention cues and the global attention mechanism for saliency detection. To improve the boundary localization of salient objects, Huang \textit{et al.}~\cite{huang2021semantic} proposed a semantic-guided attention refinement network for RSI SOD. 
As far as we know, there is no weakly-supervised SOD study for remote sensing images. This paper proposes the first weakly-supervised SOD model, which learns the saliency of RSI from the scribble annotations.

\begin{figure*}[hbt]
	\centering
	\begin{overpic}[width=1\linewidth]{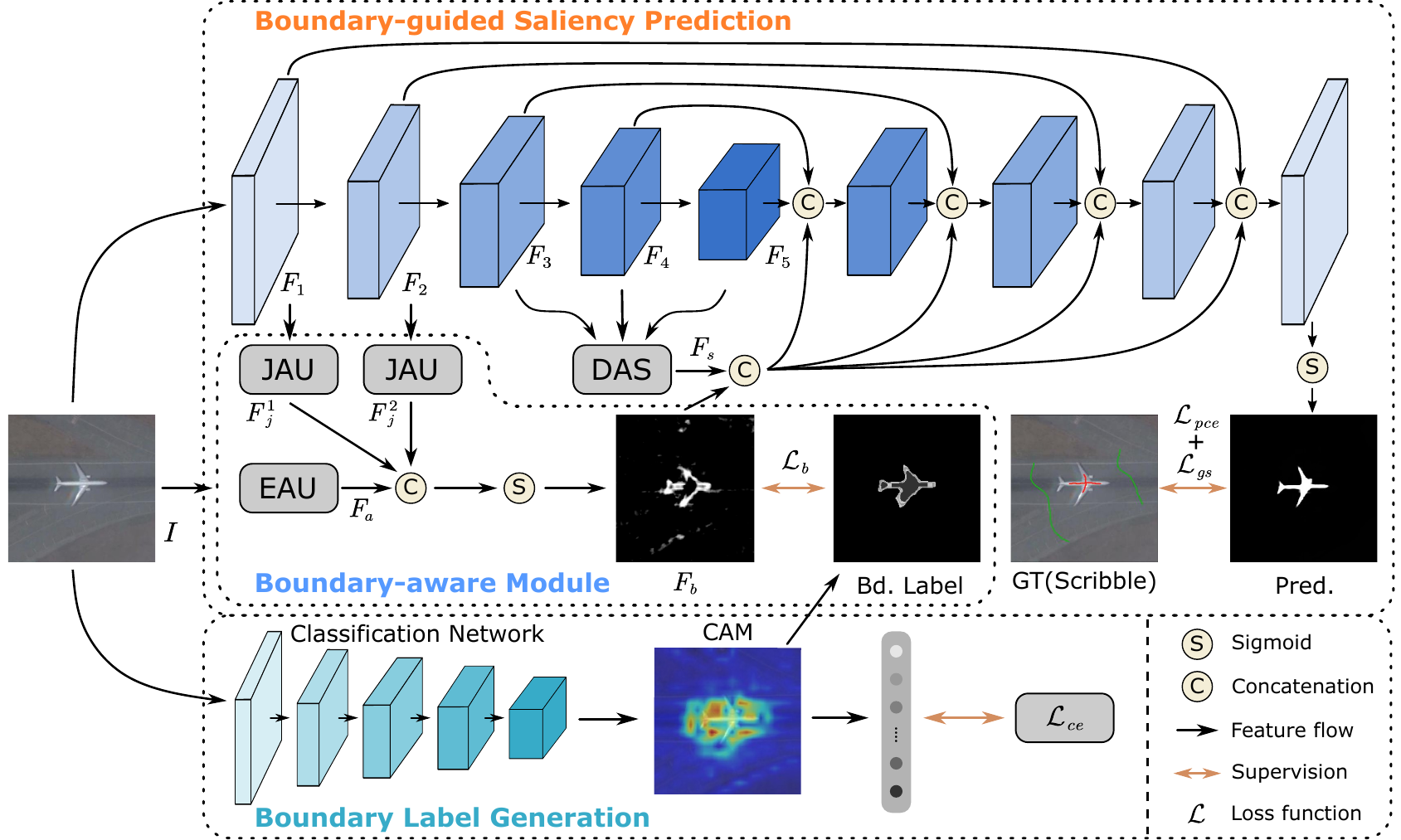}
	\end{overpic}
	\caption{
		The overall structure of the proposed \textit{SBA-Net}. It mainly contains three components, \textit{i.e.}, boundary-guided saliency prediction network, boundary-aware module (BAM), and boundary label generation (BLG) module. 
		The saliency prediction network exploits an encoder-decoder network for saliency prediction, which uses scribble annotations as supervision signals. The BAM is designed to learn the boundary semantics ($F_b$) from the low-level features ($F_1$ and $F_2$) and the input image ($I$), and is integrated with the initial saliency map ($F_s$), generated by the dense aggregation strategy (DAS) from high-level features ($F_3, F_4$ and $F_5$), to guide the saliency prediction in the decoder network. The BLG module is adopted to generate reliable boundary (pseudo) labels of objects (Bd. label) as the supervision of BAM, which is comprised of a classification network with the class activation map (CAM). The BLG module is trained with image-level class labels by cross-entropy loss $\mathcal{L}_{ce}$. JAU: joint attention unit; EAU: edge auxiliary unit.
	}\label{fig:overll}
\end{figure*}

\section{Proposed Method}

\subsection{Framework Overview}

The overall architecture of the proposed scribble-based boundary-aware SOD network for RSI is shown in Fig.~\ref{fig:overll}, which consists of boundary-guided saliency prediction network, boundary-aware module (BAM) and boundary label generation (BLG) module. 
The saliency prediction network is a basic encoder-decoder network to generate the saliency map under the supervision of scribble annotations. To produce high-quality saliency maps, we integrate the high-level features ($F_3, F_4$ and $F_5$), which contain the localization semantics of salient objects, by dense aggregation strategy (DAS) to generate the initial saliency map, and then adopt it to guide the decoder network, which gradually restores the entire salient object regions. 
From our observations, the sparse scribble annotations convey limited information, so it is difficult to infer the structure or boundary of salient objects. With this in mind, we design the BAM which exploits low-level features ($F_1$ and $F_2$) and the image to excavate object boundary cue, which later is integrated with DAS output to further guide and refine the saliency map in the decoder network. 
To learn object boundary semantics in BAM, in the BLG module, we adopt a classification network with class activation maps (CAM) to generate the object-boundary (pseudo) labels for explicit supervision of BAM. Here we exploit a sliding window strategy to filter the object location maps produced by CAM and thus obtain some reliable boundary labels. 
Noted that, we are not simply extracting the edge prior of the image to improve the saliency detection, but exploring the edge information related to the boundary of objects; that is, the specific edge cues, to guide the detection of salient objects.

\begin{figure}[bt]
	\centering
	\begin{overpic}[width=0.7\linewidth]{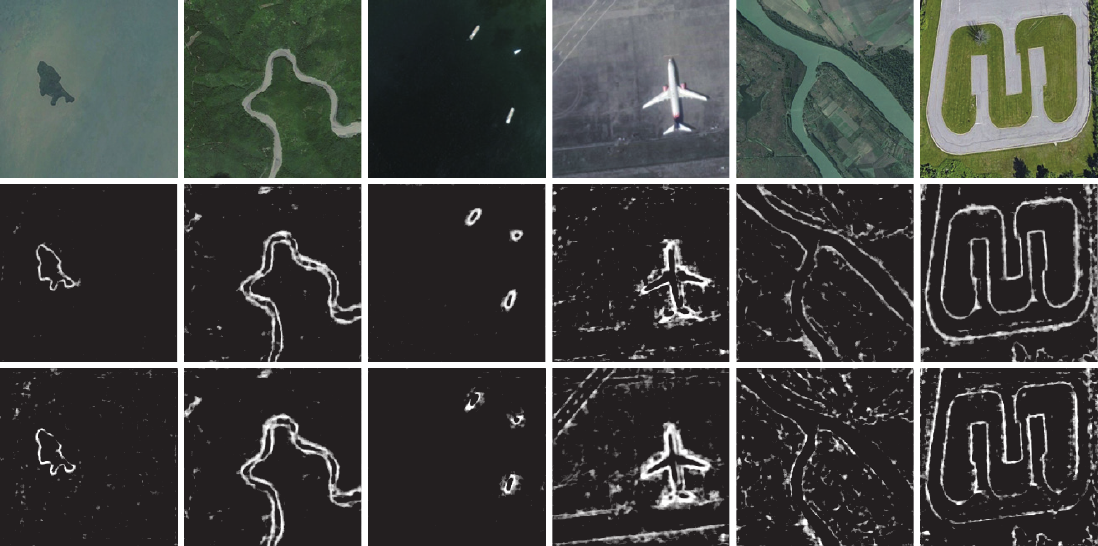}
	\end{overpic}
	\caption{Visual results of BAM. From top to bottom, they are input images, boundary detection by BAM, and boundary detection without EAU. It can be seen taht EAU is able to complement numerous edge information, thereby promoting the continuous and complete edge structures of objects.}
	\label{fig:boundary}
\end{figure}

\subsection{Boundary-aware Module}

As is known to all, sparse scribble annotations contain limited information and are difficult to train the model effectively for the inference of object structures, thus resulting in poor boundary localization in saliency maps. One promising solution is to force the model to learn more object structure semantics. Thus, some works~\cite{feng2019attentive,zhao2019egnet,zhang2020dense} propose to adopt edge detection as an auxiliary task in the SOD framework to enhance the model's ability to perceive object structure. Inspired by these works, here we design a boundary-aware module (BAM) to explore the object boundary semantics and encourage the model to produce saliency features with rich structure information.

\textbf{Module structure.} 
As shown in Fig.~\ref{fig:overll}, we build a joint attention unit (JAU) to learn the edge cues from the lateral output features ($F_1$ and $F_2$) of the backbone network, respectively. To explore more edge prior information, we also design an edge auxiliary unit (EAU) to extract the edge features ($F_a$) directly from the input image. Then, these edge features are concatenated and then activated by the sigmoid function to generate boundary feature maps, which can be calculated as:
\begin{equation}\label{equ1}
	F_b = \sigma\left(Conv\left( Concat\left( JA\left(F_1\right), JA\left(F_2\right), F_{a}\right) \right) \right),
\end{equation}
where $JA$ represents the joint attention unit, $Concat\left(\cdot \right)$ and $Conv\left(\cdot \right) $  denote concatenation and convolution operations respectively, and $\sigma$ is the sigmoid function. The BAM is explicitly supervised by the object boundary labels obtained from the boundary label generation (BLG) module introduced in Sec.~\ref{blg}. With the designed supervision strategy, the BAM can learn features that highlight the object boundary-related edges. The boundary detection result generated by BAM is shown in Fig.~\ref{fig:boundary}. It can be seen that the proposed BAM can capture rich and continuous boundary-related edges of the objects, which is beneficial for the boundary localization of salient objects.

\begin{figure}[bt]
	\centering
	\begin{overpic}[width=0.7\linewidth]{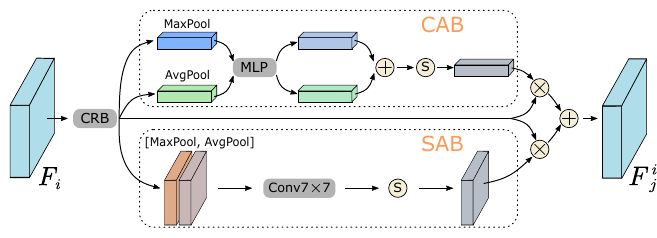}
	\end{overpic}
	\caption{The structure of joint attention unit (JAU). The input features ($F_i, i\in \{1,2\}$) are first fed into a small network with convolution, ReLU, and batch normalization layers (CRB). Then the channel attention block (CAB) and spatial attention block (SAB) are adopted to refine features. Based on element-wise multiplication ($\otimes$) and element-wise summation ($\oplus$), we obtain the edge feature maps $F_j^i$. MaxPool and AvgPool are the maximum and average pooling operations, respectively. MLP denotes a multi-layer perceptual structure with a hidden layer. 
	}\label{fig:jau}
\end{figure}

\textbf{Joint attention unit.}
Remote sensing images are often with complex backgrounds; thus, it is necessary to assign different weights to different spatial positions and channels of feature maps, so that the network can pay attention to critical features and suppress background and non-salient object features. 
Therefore, we introduce the JAU, shown in Fig.~\ref{fig:jau}, which combines channel and spatial attention mechanisms~\cite{fu2019dual} to achieve feature refinement and edge detection, which can be expressed as follows:
\begin{equation}\label{equ2}
	\begin{aligned}
		F^{'}_i & = CRB_{3\times 3}\left( F_i \right), i\in \{1, 2\} \\
		F_{j}^{i} & =\left(CAB\left( F^{'}_i \right) \otimes F^{'}_i  \right) \oplus\left( SAB\left( F^{'}_i \right) \otimes F^{'}_i  \right)  \\
	\end{aligned},
\end{equation}
where $CRB_{n\times n} $ is a small network composed of $n \times n$ convolution, ReLU, and batch normalization layers, and $n$ is the kernel size of convolution. $\oplus$ is element-wise summation. $CAB$ and $SAB$ denote channel attention block and spatial attention block, respectively. $F_j^i$ is the learned edge feature maps.

\begin{figure}[bt]
	\centering
	\begin{overpic}[width=0.65\linewidth]{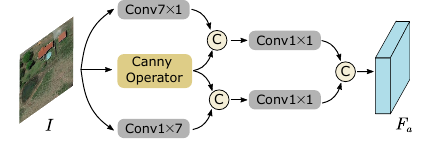}
	\end{overpic}
	\caption{
		The structure of edge auxiliary unit (EAU). The vertical and horizontal edge features of the input image are extracted through convolutinal blocks with  $7 \times1$  and $1 \times7$ convolution operations, respectively. Then they are concatenated with the edge features captured by the Canny operation to obtain rich edge features. The final edge feature map can be obtained by $1\times1$ convolution operation and concatenation.
	}\label{fig:eau}
\end{figure}

\textbf{Edge auxiliary unit.} 
Inspired by~\cite{zhang2020weakly,tian2020weakly}, we present an edge auxiliary unit to further explore edge information, which extracts edge cues directly from the input image, shown in Fig.~\ref{fig:eau}. 
Specifically, we adopt convolutional blocks with $7\times1$  and $1\times7$ convolution operations respectively to extract edge features in the vertical and horizontal directions, and then concatenate them with the edge feature obtained by the Canny operator~\cite{canny1986computational}. Finally, a $1\times1$ convolution operation is applied to adjust the channel number to generate the final features $F_a$ with rich edges. 
Although the introduction of the edge auxiliary module increases the edges of non-objects, it contributes much to the object structure and details. With the supervision of reliable boundary labels from the BLG module, more object edge cues can be learned to boost the boundary localization and the saliency map with complete structure. As shown in Fig.~\ref{fig:boundary}, EAU can complement a large amount of edge information, thereby facilitating the continuous and complete edge structures of objects, such as columns 1, 3, 5, and 6.

\subsection{Boundary Label Generation}\label{blg}

In weakly supervised detection, edge prior is often used as the supervision source for model training to enhance the structure information of predictions~\cite{wang2019boundary,zhang2020weakly,wei2021scribble}. However, the commonly used edge detection method~\cite{xie2015holistically} is generally unable to distinguish the edges as object class or background class, and thus it is easy to introduce background noises to contaminate model training.
To force the model to pay more attention to the boundary of objects rather than the edge of the whole scene, we introduce the class activation maps (CAM), which obtain the class localization maps from a classification network under the supervision of image-level class annotations, to generate reliable boundary labels. The CAM architecture can activate the most discriminative object regions and distinguish them as foreground if they contribute to the classification score.
Specifically, this is usually implemented by integrating a global average pooling (GAP) after a fully convolutional network. However, this GAP operation takes the average value of all pixels in the class localization map as the classification score; that is, the non-object regions will also be considered. 
Therefore, we introduce the attention pooling operation~\cite{chen2020weakly}, which can dynamically assign weights to pixels. 
Let $M^c_i$ represent the activation value of class $c\in C$ in the class localization map at spatial position $i$, and $A^c_i$ denote the attention value of class $c$ in the attention map at spatial position $i$. Then, the classification score $S^c $ can be obtained by summing up the element-wise multiplication of all pixels, which can be formulated as:
\begin{equation}\label{equ3}
	S^c=\sum_{i}{\left(M^c_i \otimes A^c_i \right) },~\forall c\in C,
\end{equation}
where the attention map $A^c$ is obtained from the corresponding class localization map $M^c$ through the soft-max operation and $\otimes$ is element-wise multiplication. After model training, the attention pooling operation will be removed. For inference, the class localization map $M^c$ will be normalized to obtain the class probability $P^c$. To obtain object boundary labels, we first convert probability to a certain class label (\textit{i.e.}, background, foreground, and uncertain labels) for each pixel by a hard threshold, which is denoted as:
\begin{equation}\label{equ4}
	p_i = 
	\begin{cases}
		\arg \mathop{max}\limits_{c\in C} \left( P^c_i\right), {\rm if}\mathop{max}\limits_{c\in C}\left( P^c_i\right)>T_f \\
		~~~~~~~~0, ~~~~~~~{\rm if} \mathop{max}\limits_{c\in C}\left( P^c_i\right)<T_b  \\
		~~~~~~255, ~~~~~~otherwise,
	\end{cases}
\end{equation}
where $p_i$ denotes the pixel label at spatial position $i$. 0 and 255 represent background label and uncertain label, respectively. $T_f$ and $T_b$ are the thresholds for segmenting the foreground and background, respectively, and are set to 0.30 and 0.07 in the experiment. 
According to \cite{chen2020weakly}, if the adjacent regions of one pixel contain an approximate number of identical foreground and background pixels, it is judged to be a boundary. To implement it, we use a sliding window to calculate the number of local identical pixels and then determine whether the pixel in the center of the window belongs to the boundary or not by statistical proportion. More details can refer to \cite{chen2020weakly}. In our experiment, we set the window size to 13. After the above steps, we can obtain credible object boundary labels to supervise BAM training.

\subsection{Boundary-guided Saliency Prediction}

\begin{figure}[tb]
	\centering
	\begin{overpic}[width=0.65\linewidth]{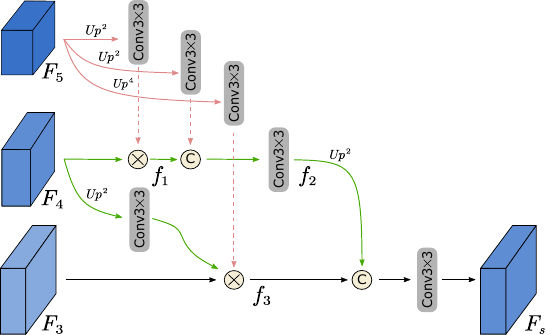}
	\end{overpic}
	\caption{
		The structure of dense aggregation strategy (DAS). This strategy starts from the high-level feature $F_5$ and densely aggregates $F_4$ and $F_3$ in a top-down manner to obtain an initial saliency map with rough boundary structure but more accurate locations.
	}\label{fig:DAS}
\end{figure}

In our saliency prediction network, we exploit an encoder-decoder structure similar to U-Net~\cite{ronneberger2015u} as our basic network architecture. If directly training this model with sparse scribble labels, it may not be easy to recover the structure and boundary of salient objects. To this end, we employ the boundary cues generated from BAM as a guidance to enable the model to predict the saliency map with complete object structure.
Specifically, we first integrate the high-level features (\textit{i.e.}, $F_3, F_4$, and $F_5$), which contains location semantics of salient objects, by a designed dense aggregation strategy (DAS) to obtain an initial saliency map with rough boundaries but accurate locations. The dense aggregation strategy is shown in Fig.~\ref{fig:DAS} and can be formulated as follows:
\begin{equation}\label{equ6}
	\begin{aligned}
		f_{1}&=Conv\left( Up^2\left( F_5 \right) \right) \otimes F_4 \\
		f_{2}&=Conv\left( Concat\left(f_1, Conv\left( Up^2\left( F_5 \right) \right) \right) \right) \\
		f_{3}&=Conv\left( Up^4\left( F_5 \right) \right) \otimes Conv\left( Up^2\left( F_4 \right) \right) \otimes F_3 \\
		F_{s}&=Conv\left( Concat\left(f_3, Conv\left( Up^2\left( f_2 \right) \right) \right) \right)
	\end{aligned},
\end{equation}
where $Up^n$ represents $n\times$ upsampling operation, and $Conv$ denotes $3\times3$ convolution layer. After that, the output initial saliency map $F_{s}$ is concatenated with the boundary feature map $F_b$ produced by BAM to guide the each upsample layer in the decoder network. Obviously, the concatenated feature can simultaneously enhance the boundary structure and positioning of salient objects. The overall structure of the salient object is restored gradually in a top-down manner under the guidance of boundary cues.

\subsection{Loss Function}

\textbf{Boundary loss.} 
The BAM outputs the prediction of object boundary and is supervised by the generated boundary labels by BLG. However, in the training process, the unevenness of the sample distribution will reduce the model perception for a boundary. Thus we adopt the relaxed boundary classification loss to calculate the cross-entropy of each part of the sample separately, which can be expressed as:
\begin{equation}\label{equ7}
	\mathcal{L}_{b}=-\sum_{i\in \Phi _{bry}}{\frac{W_i\log \left( P_i \right)}{\left| \Phi _{bry} \right|}}-\frac{1}{2}\sum_{i\in \Phi _{fb}}{\frac{\log \left( 1-P_i \right)}{\left| \Phi _{fb} \right|}},
\end{equation}
where $P_i$ is the boundary probability for pixel $i$ predicted by BAM, and $\Phi_{bry} $ and $\Phi_{fb}$ are the pixel sets from the boundary and non-boundary regions (including foreground and background) of the generated labels, respectively. Here, we add a weight $W_i $ to the boundary loss term to relax the dependence on boundary labels.

\textbf{Structure loss.} 
Considering that the expected saliency maps should be with consistent probability within objects and sharp object boundaries. Therefore, we introduce the gated structure-aware loss~\cite{zhang2020weakly} to maintain the structure of the entire object region while enhancing the smoothness. The loss function is expressed as:
\begin{equation}\label{equ8}
	\mathcal{L}_{gs}=\sum_{i}{\sum_{d\in \vec{x},\vec{y}}{\Psi}}\left( \left| \partial _dS_{i} \right|e^{-\alpha \left| \partial _d\left( G\cdot I_{i} \right) \right|} \right),
\end{equation}
where $S_i$ denotes the value of pixel $i$ in predicted saliency map, $\Psi$ is calculated as $ \Psi(S) = \sqrt{S^2+1e^{-6}} $ to avoid the square root of zero, $I_{i} $ is the image intensity of pixel $i$, $d$ is the partial derivative in the $\vec{x}$ and $\vec{y}$ directions, and $G$ is the gated for the structure-aware loss. The $\mathcal{L}_{gs}$ loss applies $L_1$ penalty to the gradients of predicted saliency map $S$ to enhance the smoothness, and $\partial I$ as weight to preserve boundaries of saliency map along image edges. More details can be see~\cite{zhang2020weakly}.

\textbf{Objective function.} 
Under the supervision of the scribble labels, we also use partial cross-entropy loss to train the model with scribble annotations, which can be defined as:
\begin{equation}\label{equ9}
	\mathcal{L} _{pce}=\sum_{i\in Scr}{-GT_i\log S_i-\left( 1-GT_i \right) \log \left( 1-S_i \right)},
\end{equation}
where $ GT $ denotes the ground truth, $ Scr $ is the pixel set of the scribble annotations. Therefore, combining boundary loss and structure loss, our final loss function is then denoted as:
\begin{equation}\label{equ10}
	\mathcal{L} _{total}=\mathcal{L} _{b}+\mathcal{L} _{gs}+\mathcal{L} _{pce}.
\end{equation}

Here, the $\mathcal{L} _{pce}$ loss uses scribble annotations as supervision to propagate the scribble pixels to the foreground regions by saliency prediction network, and $\mathcal{L} _{gs}$ loss exploits image boundary information as supervision, and $\mathcal{L}_{b}$ loss boosts the boundary localization of salient objects.

\section{Experimental Results}
\subsection{Scribble Dataset}
Due to the lack of a scribble annotation-based dataset for RSI saliency detection, we have re-labeled the existing EORSSD dataset \cite{zhang2020dense}, the largest dataset available for RSI SOD, with scribbles, namely S-EOR dataset. Compared to ORSSD~\cite{li2019nested}, the EORSSD dataset contains 2,000 optical remote sensing images with more challenging complex scenes and more diverse object attributes.

\begin{figure}[bt]
	\centering
	\begin{overpic}[width=0.7\linewidth]{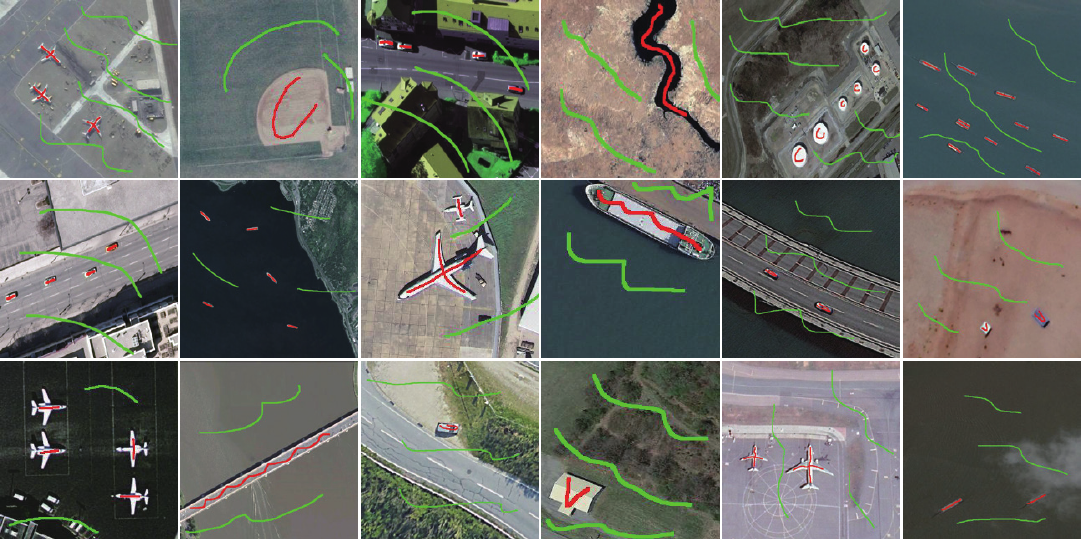}
	\end{overpic}
	\caption{
		Visualization of scribble annotations in some typical scenes, in which salient objects and background regions are labeled in red and green, respectively. The pixels of the unlabeled regions indicate unknown pixels.
	}\label{fig:ann-vis}
\end{figure}

\begin{figure}[bt]
	\centering
	\begin{overpic}[width=0.7\linewidth]{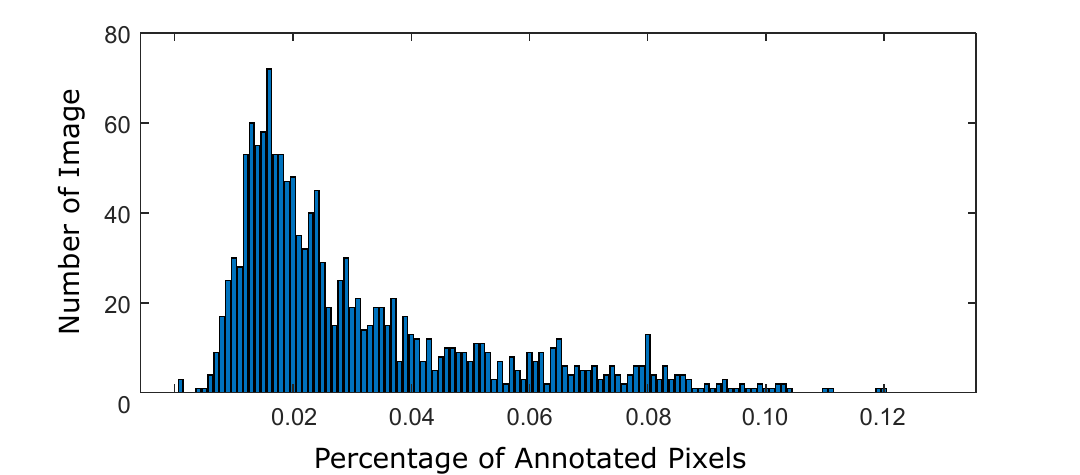}
	\end{overpic}
	\caption{
		Percentage of annotated pixels in the S-EOR dataset. The statistical percentage of marked pixels is 3.03\%, demonstrating the sparsity of annotations.
	}\label{fig:percentage}
\end{figure}

Following the annotation process in~\cite{zhang2020weakly}, our annotators distinguished and annotated the EORSSD dataset via scribbles based on the first impression of the objects in the scene and did not rely on the given ground-truth salient objects. For an image, the annotator can complete the scribble annotation in only 3$\sim$4 seconds. Fig.~\ref{fig:ann-vis} provides some examples of scribble annotations in typical scenarios. Fig.~\ref{fig:percentage} shows the statistics of the percentage of annotated pixels in the entire S-EOR dataset. The average percentage of labeled pixels (both foreground and background) is 3.03\%, and other pixels are regarded as uncertain pixels, demonstrating the sparsity of annotations and the weakness of supervision signals that can be used for model training.
Note that to generate the boundary labels of objects by BLG module, we also provide the class labels of each sample (image-level annotations) for supervision of BLG.
In addition, in the training stage, we only use scribble annotations as the supervision signal for the proposed model. In the testing stage, the pixel-level saliency map is inferred directly from the input image. We evaluate our method on two widely used RSI SOD benchmark datasets, \textit{i.e.}, ORSSD \cite{li2019nested} and EORSSD \cite{zhang2020dense}.

\subsection{Evaluation Metrics}
For a comprehensive quantitative comparison, we adopt six widely used SOD evaluation metrics, including mean absolute error $\left(MAE, \mathcal{M}\right)$ \cite{perazzi2012saliency}, average F-measure $\left( F_{avg}\right)  $, maximum F-measure $\left( F_{max}\right)  $ \cite{achanta2009frequency}, average E-measure $ \left( E_{avg} \right) $, maximum E-measure $ \left( E_{max} \right) $ \cite{fan2018enhanced}, and S-measure  $\left(S_{\xi}\right)$ \cite{fan2017structure}. In addition, the precision-recall (PR) curves and the F-measure curves are also used to compare with other state-of-the-art (SOTA) models. The details of these evaluation metrics are as follows:

\textbf{\emph{MAE}} \bm{$\left(\mathcal{M}\right)$} evaluates the average per-pixel difference between the predicted saliency map and its ground-truth map after normalization processing, which is defined as follows:
\begin{equation}\label{equ11}
	\mathcal{M}=\frac{1}{H \times W} \sum_{i=1}^{H} \sum_{j=1}^{W}|P(i, j)-GT(i, j)|,
\end{equation}
where $ W $ and $ H $ are the width and height of evaluated image, respectively, and $i$ and $j$ are the horizontal and vertical coordinates of the image pixels, respectively.

\textbf{F-measure} \bm{$\left(F_{\beta}\right)$} is used to comprehensively evaluate the precision and recall, which is calculated as follows:
\begin{equation}\label{equ12}
	F_{\beta}=\frac{\left(1+\beta^{2}\right) \text { Precision } \times \text { Recall }}{\beta^{2} \times \text { Precision }+\text { Recall }}.
\end{equation}
Here we set $\beta^{2}$ to 0.3 and report the average F-measure $\left( F_{avg}\right)  $, maximum F-measure $\left( F_{max}\right) $ for each dataset.

\textbf{E-measure} \bm{$\left(E_{\phi}\right)$} is a cognitive vision-based metric that combines local pixel values with image-level average values to evaluate the similarity with GT images jointly. The E-measure is calculated as follows:
\begin{equation}\label{equ13}
	E_{\phi}=\frac{1}{W\times H}\sum_{i=1}^{W}\sum_{i=1}^{H}\theta\left(\phi\right),
\end{equation}
where $ \theta\left(\phi\right) $ is the enhanced alignment matrix \cite{fan2018enhanced}. We also report the average E-measure $ \left( E_{avg} \right) $ and maximum E-measure $ \left( E_{max} \right) $ for each dataset.

\textbf{S-measure} \bm{$\left(S_{\xi}\right)$} considers the structural similarity of object perception ($S_{o}$) and range ($S_{r}$) perception from the perspective of image structure information. S-measure can be calculated as:
\begin{equation}\label{equ14}
	S_{\xi}=\alpha \times S_{o}+\left(1-\alpha\right)\times S_{r},
\end{equation}
where $\alpha$ is the trade-off parameter and set to 0.5.

\subsection{Implementation Details}
Our model is implemented based on the PyTorch framework~\cite{paszke2017automatic}. 
For the boundary label generation module, we adopt ResNet-50~\cite{he2016deep} as the backbone network, then replace the last fully connected layer with a $1\times 1$ convolution layer and halve the stride of the last stage to preserve more spatial information. We first train the BLG module using the image-level class annotations with cross-entropy classification loss ($\mathcal{L} _{ce}$), to generate object boundary (pseudo) labels.
Then in the boundary-guided saliency prediction network, we use VGG-16~\cite{simonyan2014very} as the backbone network. 
For model training under the supervision of scribbles, we adopt 1,400 newly labeled images in the S-EOR dataset as training data. Then we use some general image augmentation strategies, such as rotation, flipping, and random cropping operations. 
The input image is resized to $352 \times352$. The Adam algorithm is used to optimize the model parameters with a learning rate of 1e-4. The whole training takes almost 6 hours on an NVIDIA GeForce RTX 3090 GPU for 80 epochs with the batch size 4.

\subsection{Comparison with State-of-the-arts}
We compare our method with 15 SOTA salient object detection models from both quantitative and qualitative aspects, including 8 fully-supervised saliency detection methods (\textit{i.e.}, R3Net \cite{deng2018r3net}, DSS \cite{hou2019deeply}, RADF \cite{hu2018recurrently}, PoolNet\cite{liu2019simple}, PFA \cite{zhao2019pyramid}, EGNet \cite{zhao2019egnet}, MINet \cite{pang2020multi}, and GateNet \cite{zhao2020suppress}), 3 fully-supervised methods designed specifically for RSI SOD (\textit{i.e.}, LVNet \cite{li2019nested}, DAFNet \cite{zhang2020dense} and SARNet \cite{huang2021semantic}), and 4 weakly-supervised/unsupervised methods for SOD (\textit{i.e.}, SVF \cite{zhang2017supervision}, MSW \cite{zeng2019multi}, WSSA \cite{zhang2020weakly}, and SCWS \cite{yu2021structure}).

\textbf{Quantitative comparison.} 
Tab.~\ref{tab:qc} shows the comparison results of our method and other competing methods under the six evaluation indicators on two datasets. 
From these results, it can be seen that the proposed method consistently outperforms the other weakly-supervised or unsupervised methods. 
Noted that, the recently proposed WSSA~\cite{zhang2020weakly} also introduces the edge prior, implemented by an auxiliary edge detection network, to enhance saliency prediction from scribble annotations, and achieves the second-best performance. In contrast, this model can not distinguish the edges of object boundaries and thus reduce the detection performance.
Thanks to our well-designed BAM and BLG modules, the proposed model is able to focus more on the object boundary while accurately locating the salient object, thereby predicting a saliency map with a more complete structure. We can see that, our method obtains much higher F-measure and E-measure scores, which show that our prediction results have higher spatial consistency and visual integrity. 
Besides, our method also achieves comparable or superior performance when compared with other fully-supervised models, such as EGNet~\cite{zhao2019egnet}, MINet~\cite{pang2020multi}, and GateNet~\cite{zhao2020suppress}, which are originally designed for natural images, and LVNet~\cite{li2019nested} which is originally designed for remote sensing images.

\begin{figure*}[thp!]
	\centering
	\begin{overpic}[width=1\linewidth]{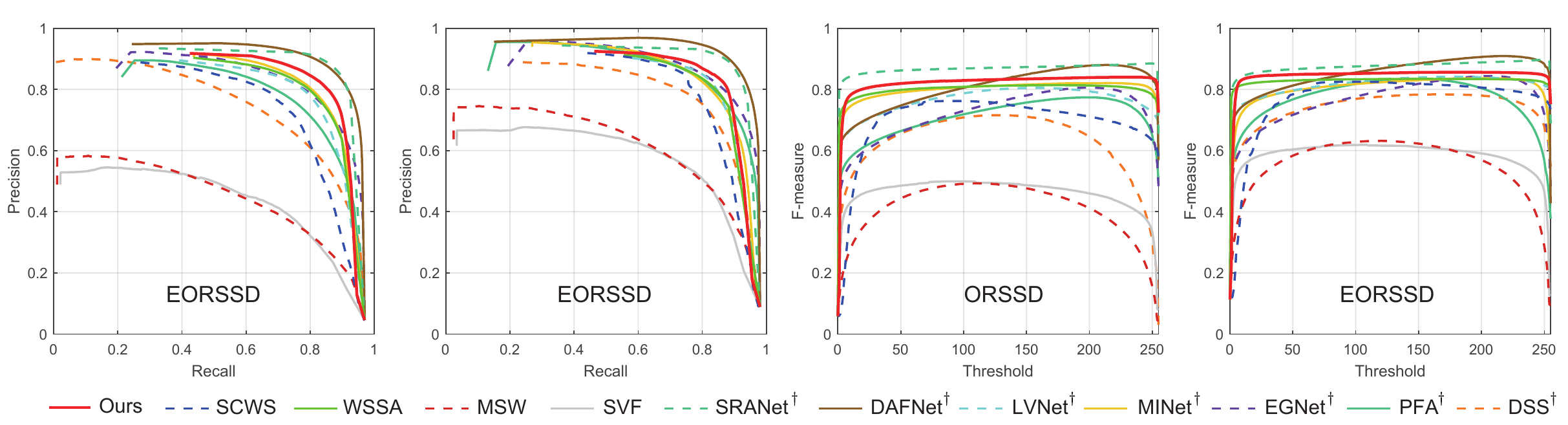}
	\end{overpic}
	\caption{
		Comparison of Precision-Recall curves and F-measure curves of 12 SOTA methods on two optical RSI datasets. ``$\dagger$'' denotes the fully supervised saliency detection methods. Our method is superior to all compared methods except for two recently proposed fully-supervised SOD models for RSI, \textit{i.e.}, SARNet \cite{huang2021semantic} and DAFNet \cite{zhang2020dense}. Please zoom in for a better view.
	}\label{fig:curve}
\end{figure*}

Fig.~\ref{fig:curve} shows the PR curves and F-measure curves of the proposed method and other SOTAs on two datasets. In addition to the two fully-supervised RSI SOD models, \textit{i.e.}, DAFNet~\cite{zhang2020dense} and SARNet~\cite{huang2021semantic}, the PR curves of our method (marked in red) are closer to the point (1,1) as a whole, which indicates that our method can detect more foreground pixels. At the same time, the F-measure curves of our method have a higher vertical axis as a whole, which reflects that the prediction results of our method have better spatial consistency and coherent details. Moreover, we can observe that our method is superior to other SOD models under different thresholds, which also demonstrates the robustness of the proposed method.

\begin{table*}[]
	\centering
	\caption{Quantitative comparison with 15 SOTA methods on two benchmark datasets, including eight fully supervised SOD methods for NSI, three fully supervised SOD methods designed for RSI, and four weakly supervised or unsupervised SOD methods. \textbf{Sup.} and \textbf{Bb.} denote the type of supervision and backbone, respectively. \textbf{F}: fully supervised. \textbf{Un}: unsupervised. \textbf{M}: multi-source weak supervision. \textbf{S}: scribble-based weak supervision. $\uparrow$ / $\downarrow$ denote the larger/smaller is better, respectively.} \vspace{2mm}
	\label{tab:qc}
	\scalebox{0.52}{
	\begin{tabular}{lcccccccc|cccccc}
		\toprule
		\multirow{2}{*}{\textbf{Models}} & \multirow{2}{*}{\textbf{Sup.}} &\multirow{2}{*}{\textbf{Bb.}} & \multicolumn{6}{c}{\textbf{ORSSD}\cite{li2019nested}}  & \multicolumn{6}{c}{\textbf{EORSSD}\cite{zhang2020dense}} \\
		\cmidrule[0.05em](lr){4-9} \cmidrule[0.05em](lr){10-15}
		&                      & & $ S_{\xi}\uparrow $ & $ E_{avg}\uparrow $ & $ E_{max}\uparrow $ & $ F_{avg}\uparrow $ & $ F_{max}\uparrow $ & $ \mathcal{M}\downarrow $   & $ S_{\xi}\uparrow $ & $ E_{avg}\uparrow $ & $ E_{max}\uparrow $ & $ F_{avg}\uparrow $ & $ F_{max}\uparrow $ & $ \mathcal{M}\downarrow $   \\
		\midrule
		R3Net \cite{deng2018r3net}      & F     &  ResNeXt-101      & 0.814    & 0.868  & 0.891 & 0.738  & 0.738 & 0.040 & 0.819    & 0.831  & 0.950 & 0.632  & 0.752 & 0.017 \\
		DSS \cite{hou2019deeply}        & F      & VGG-16       & 0.826    & 0.836  & 0.886 & 0.696  & 0.747 & 0.036 & 0.787    & 0.764  & 0.920 & 0.582  & 0.687 & 0.019 \\
		RADF \cite{hu2018recurrently}   & F       & VGG-16       & 0.826    & 0.830  & 0.913 & 0.686  & 0.762 & 0.038 & 0.819    & 0.859  & 0.916 & 0.661  & 0.747 & 0.017 \\
		PoolNet  \cite{liu2019simple}   & F     & ResNet-50          & 0.840    & 0.865  & 0.943 & 0.700  & 0.771 & 0.036 & 0.822    & 0.821  & 0.932 & 0.643  & 0.758 & 0.021 \\
		PFA  \cite{zhao2019pyramid}     & F      & VGG-16       & 0.861    & 0.855  & 0.952 & 0.731  & 0.813 & 0.024 & 0.836    & 0.866  & 0.929 & 0.679  & 0.748 & 0.016 \\
		EGNet  \cite{zhao2019egnet}     & F      &  VGG-16       & 0.872    & 0.901  & 0.973 & 0.750  & 0.833 & 0.022 & 0.860    & 0.877  & 0.957 & 0.697  & 0.788 & 0.011 \\
		MINet  \cite{pang2020multi}     & F      &ResNet-50       & 0.849    & 0.894  & 0.929 & 0.779  & 0.812 & 0.028 & 0.858    & 0.915  & 0.937 & 0.772  & 0.791 & 0.013 \\
		GateNet   \cite{zhao2020suppress}   & F    & ResNet-50       & 0.893    & 0.927  & 0.955 & 0.827  & 0.860 & 0.015 & 0.880    & 0.904  & 0.946 & 0.770  & 0.812 & 0.011 \\
		\midrule
		LVNet   \cite{li2019nested}      & F        & -       & 0.882    & 0.926  & 0.946 & 0.800  & 0.826 & 0.021 & 0.864    & 0.883  & 0.928 & 0.736  & 0.782 & 0.015 \\
		DAFNet   \cite{zhang2020dense}       & F  & Res2Net-50       & 0.919    & \textbf{0.954}  & \textbf{0.982} & 0.844  & \textbf{0.900 }& \textbf{0.011} & 0.918    & 0.938  & \textbf{0.982} & 0.798  & \textbf{0.873} & \textbf{0.005} \\
		SARNet   \cite{huang2021semantic}    & F   & VGG-16      & \textbf{0.913}    & 0.948  & 0.956 & \textbf{0.862}  & 0.885 & 0.019 & \textbf{0.924}    & \textbf{0.955}  & 0.962 & \textbf{0.854 } & 0.872 & 0.010 \\
		\midrule
		SVF    \cite{zhang2017supervision}   & Un   & VGG-16       & 0.713    & 0.724  & 0.780 & 0.521  & 0.577 & 0.075 & 0.656    & 0.631  & 0.703 & 0.377  & 0.440 & 0.058 \\
		MSW     \cite{zeng2019multi}         & M  & DenseNet-169       & 0.727    & 0.683  & 0.817 & 0.483  & 0.590 & 0.069 & 0.660    & 0.609  & 0.747 & 0.342  & 0.442 & 0.060 \\
		WSSA  \cite{zhang2020weakly}   & S & VGG-16       & 0.848    & 0.908  & 0.908 & 0.806  & 0.814 & 0.027 & 0.852    & 0.901  & 0.925 & 0.771  & 0.778 & 0.015 \\
		SCWS  \cite{yu2021structure}      & S  & ResNet-50   & 0.725    & 0.799  & 0.886 & 0.722  & 0.801 & 0.096 & 0.654    & 0.737  & 0.836 & 0.626  & 0.721 & 0.094 \\
		
		\textbf{Ours}                                   &  S & VGG-16      & \textbf{0.862}    & \textbf{0.919}  & \textbf{0.939} & \textbf{0.825}  & \textbf{0.837} & \textbf{0.026} & \textbf{0.861}    & \textbf{0.922}  & \textbf{0.938} & \textbf{0.798}  & \textbf{0.816} & \textbf{0.014 }\\
		\bottomrule
	\end{tabular}
	}
\end{table*}

\begin{figure*}[thp!]
	\centering
	\begin{overpic}[width=1\linewidth]{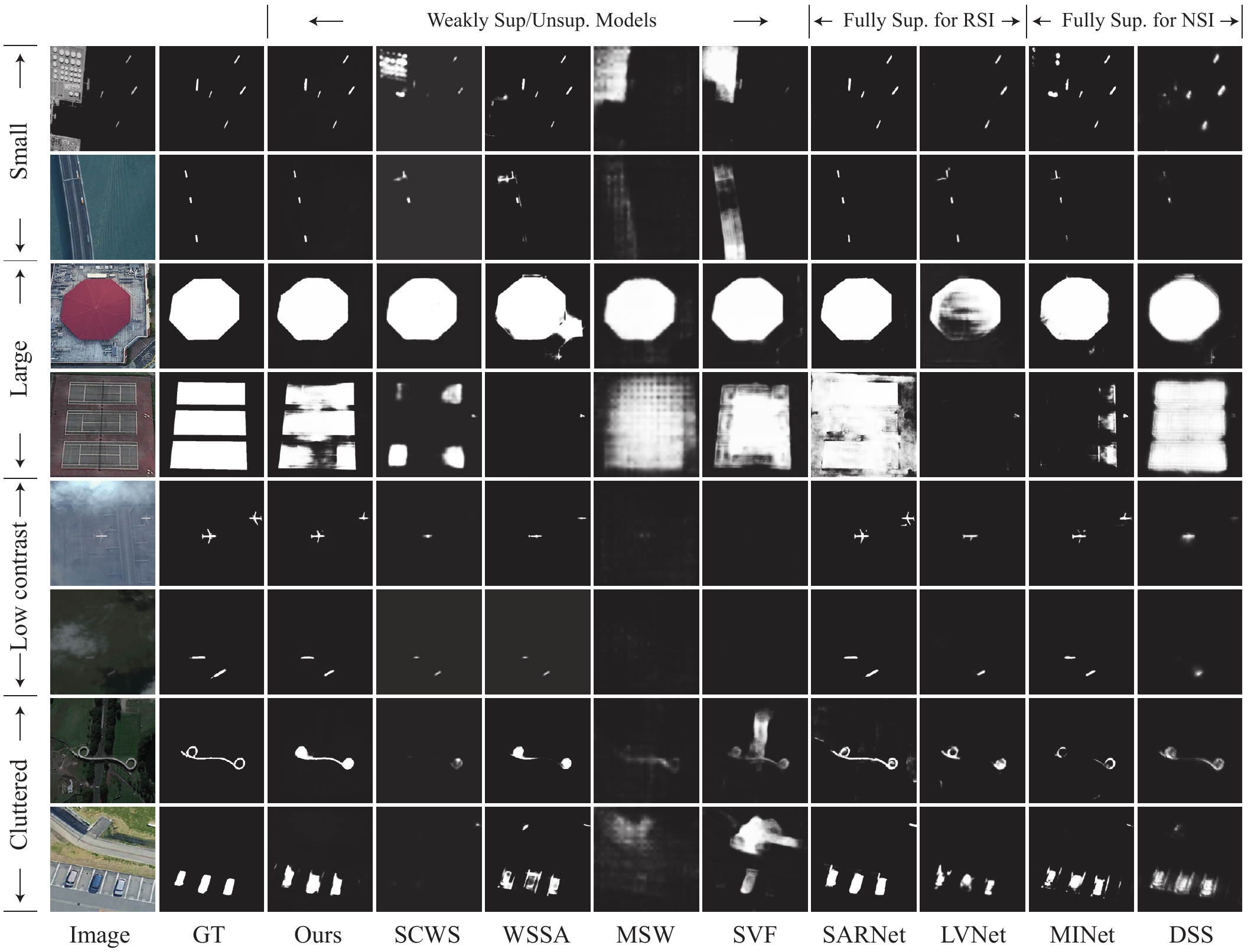}
	\end{overpic}
	\caption{Visual comparison of the proposed method with other competitors under various challenging remote sensing scenarios, including small objects, large objects, low contrast, and cluttered backgrounds. It can be seen that, our method achieves superior visual performance to other weakly-supervised/unsupervised and fully-supervised methods. Please zoom in for a better view.
	}
	\label{fig:main-VC}
\end{figure*}

\textbf{Qualitative comparison.} 
Fig.~\ref{fig:main-VC} shows the visual comparison of various challenging scenes, including scenes with small objects, large objects, low contrast between salient objects and backgrounds, and cluttered scenes. 
We provide the visual comparison of four weakly supervised/unsupervised SOD models (\textit{i.e.}, SVF~\cite{zhang2017supervision}, MSW~\cite{zeng2019multi}, WSSA~\cite{zhang2020weakly}, and SCWS~\cite{yu2021structure}), two fully-supervised SOD models designed specifically for RSI (\textit{i.e.}, LVNet~\cite{li2019nested} and SARNet~\cite{huang2021semantic}), and two fully-supervised models from natural images (\textit{i.e.}, DSS~\cite{hou2019deeply} and MINet~\cite{pang2020multi}).
It can be seen that although our method only uses scribble annotations as supervision signals, it can still accurately capture the location of the object and maintain the internal smoothness and sharp boundary of the salient object. 

\textit{Scale change.} 
The stable perception of multi-scale objects has always been a great challenge for deep detection models. From the 1st row to the 4th row in Fig.~\ref{fig:main-VC}, we can see that the proposed model can adapt to various scenes and  accurately discover and locate salient objects with different scales and shapes.
In the small object scene, the weakly-supervised and unsupervised deep learning models fail to detect ships and vehicles, and misdetect some non-salient objects. In the large object scenes, most methods provide saliency maps with poor performance, such as the badminton court scene.

\textit{Low contrast.} 
In the low-contrast environment, the objects are easily submerged in the background, due to the low illumination or similar textures, especially for small objects. As can be seen in Fig.~\ref{fig:main-VC} (rows 5 and 6), most methods cannot detect salient objects in these cases, especially weakly supervised/non-supervised methods. On the contrary, the proposed method can predict sharp saliency maps.

\textit{Cluttered background.} 
The ability to capture objects from a cluttered background is critical to the detector. As shown in Fig.~\ref{fig:main-VC} (rows 7 and 8), although the proposed model cannot get perfect prediction maps, we can accurately locate the object and obtain internally consistent results compared with other methods. Overall, our method achieves superior visual performance compared to other comparison methods.

\begin{table*}[]
	\centering
	\renewcommand{\arraystretch}{1.2}
	\renewcommand{\tabcolsep}{1.1mm}
	\caption{The ablation study of different components for the proposed model on the ORSSD and EORSSD datasets, in which ``BS'' denotes the baseline, ``DAS'' denotes dense aggregation strategy. ``SC'', ``CA'', and ``JAU'' refer to simple concatenation operation, channel attention module, and joint attention unit to explore the edge information from the lateral output features $F_1$  and $F_2$, respectively. ``EAU'' is the edge auxiliary unit.} \vspace{2mm}
	\label{tab:as}
	\scalebox{0.6}{
	\begin{tabular}{ccccccc|cccccc|cccccc}		
		\toprule
		\multirow{2}{*}{\textbf{No.}} & \multicolumn{6}{c}{\textbf{Settings}} & \multicolumn{6}{c}{\textbf{ORSSD}\cite{li2019nested}}                          & \multicolumn{6}{c}{\textbf{EORSSD}\cite{zhang2020dense}}                         \\
		\cmidrule[0.05em](lr){2-7} \cmidrule[0.05em](lr){8-13} \cmidrule[0.05em](lr){14-19}
		& BS   & DAS   & SC & CA & JAU  & EAU   & $ S_{\xi}\uparrow $ & $ E_{avg}\uparrow $ & $ E_{max}\uparrow $ & $ F_{avg}\uparrow $ & $ F_{max}\uparrow $ & $ \mathcal{M}\downarrow $   & $ S_{\xi}\uparrow $ & $ E_{avg}\uparrow $ & $ E_{max}\uparrow $ & $ F_{avg}\uparrow $ & $ F_{max}\uparrow $ & $ \mathcal{M}\downarrow $   \\
		\midrule
		1 & $ \checkmark $ &   &   &   &   &   & 0.828 & 0.890 & 0.914 & 0.789 & 0.804 & 0.042 & 0.815 & 0.901 & 0.925 & 0.783 & 0.803 & 0.031 \\
		2 & $ \checkmark $ & $ \checkmark $ &   &   &   &   & 0.837 & 0.896 & 0.912 & 0.789 & 0.807 & 0.036 & 0.812 & 0.900 & 0.924 & 0.775 & 0.806 & 0.029 \\
		3 & $ \checkmark $ & $ \checkmark $ & $ \checkmark $ &   &   &   & 0.842 & 0.886 & 0.911 & 0.790 & 0.812 & 0.034 & 0.823 & 0.901 & 0.926 & 0.778 & 0.806 & 0.025 \\
		4 & $ \checkmark $ & $ \checkmark $ &   & $ \checkmark $ &   &   & 0.849 & 0.895 & 0.917 & 0.791 & 0.819 & 0.034 & 0.836 & 0.903 & 0.929 & 0.783 & 0.809 & 0.023 \\
		5 & $ \checkmark $ & $ \checkmark $ &   &   & $ \checkmark $ &   & 0.857 & 0.910 & 0.931 & 0.817 & 0.829 & 0.029 & 0.852 & 0.914 & 0.932 & 0.791 & 0.812 & 0.017 \\
		6 & $ \checkmark $ & $ \checkmark $ &   &   & $ \checkmark $ & $ \checkmark $ & 0.862 & 0.919 & 0.939 & 0.825 & 0.837 & 0.026 & 0.861 & 0.922 & 0.938 & 0.798 & 0.816 & 0.014 \\
	
		\bottomrule
	\end{tabular}
	}
\end{table*}

\subsection{Ablation Study}
In this section, we conduct different ablation studies to analyze the contribution of each component to the overall network. The adopted baseline model (BS) is similar to FCN8S~\cite{long2015fully}. Then we integrate each designed module to verify its effectiveness. The effect of each component on two public datasets under six evaluation metrics is shown in Tab.~\ref{tab:as}.

\textbf{Effectiveness of Dense Aggregation Strategy (DAS)}. 
Based on the baseline model, we introduce the DAS to aggregate the three high-level features to generate the initial saliency map, which is then integrated into the decoder as guidance for final saliency prediction. As shown in Tab.~\ref{tab:as} (No.~2), with DAS, the MAE score decreased by 14.3\% and 6.5\% on ORSSD and EORSSD datasets respectively, and other metrics are on par with the baseline model.

\textbf{Effectiveness of Joint Attention Unit (JAU)}. 
In our model, the boundary cue is introduced to integrate with DAS output to guide the saliency prediction. JAU is designed to explore the edge semantics from two low-level features. To test the effectiveness of the joint attention unit, we introduce the simple concatenation operation (SC) and the channel attention module (CA) as a comparison. As reported in Tab.~\ref{tab:as} (No.~3$\sim$5), we can see the boundary guidance has significantly improved the detection performance no matter what type of operation is used. 
Taking EORSSD dataset as an example, compared with no boundary guidance (\textit{i.e.}, No.~2), the performance gains of JAU are (6.0\%, 2.4\%, 1.5\%, 3.0\%, 1.2\%, 41.2\%) for the metrics $S_{\xi}$, $E_{avg}$, $E_{max}$, $F_{avg}$, $F_{max}$ and $\mathcal{M}$, respectively. 
Compared with SC, the performance is improved by (4.6\%, 2.3\%, 1.3\%, 2.6\%, 1.2\%, 32.0\%) for the metrics $S_{\xi}$, $E_{avg}$, $E_{max}$, $F_{avg}$, $F_{max}$ and $\mathcal{M}$, respectively.
Then compared with CA, the performance increases are (3.0\%, 2.1\%, 1.0\%, 1.9\%, 0.9\%, 26.1\%), respectively. Obviously, the proposed JAU module significantly improves the performance of the model for saliency detection.

\textbf{Effectiveness of Edge Auxiliary Unit (EAU)}. 
The edge auxiliary unit is introduced to supplement more detailed edge information for the boundary exploration of the model. As shown in Tab.~\ref{tab:as} (No.~6), with the adding of EAU, the detection performance is further improved under all metrics on both two datasets. Compared with JAU, the performance gains are (0.6\%, 1.0\%, 0.9\%, 1.0\%, 1.0\%, 10.3\%) and (1.1\%, 0.9\%, 0.6\%, 0.9\%, 0.5\%, 17.6\%) for the metrics $S_{\xi}$, $E_{avg}$, $E_{max}$, $F_{avg}$, $F_{max}$ and $\mathcal{M}$, respectively. 

To sum up, the proposed model achieves excellent performance. Specifically, compared with BS,  the $S_{\xi}$ gains 4.1\% and 5.6\%, the $E_{avg}$ gains 3.3\% and 2.3\%, the $E_{max}$ gains 2.7\% and 1.4\%, the $F_{avg}$ increases 4.6\% and 1.9\%,  $F_{max}$ increases 4.1\% and 1.6\%, and the $\mathcal{M}$ achieves 38.1\% and 54.8\% decreases, on the ORSSD and EORSSD datasets, respectively.

\section{Conclusion}
This paper focuses on studying weakly-supervised salient object detection for remote sensing images. Firstly, we construct the first weakly-supervised remote sensing dataset by relabelling the existing dataset with scribble annotations. Then, we propose a novel weakly-supervised SOD model, \textit{i.e.}, a scribble-based boundary-aware network (SBA-Net), to learn the saliency of remote sensing images from scribbles. Specifically, we design a boundary label generation module, consisting of a classification network with the class activation map, to generate reliable boundary (pseudo) labels of the objects to supervise the boundary-aware module. To produce the high-quality saliency maps, the boundary-aware module is presented to distill the boundary semantic from low-level features and the input image to enforce the model focus more on salient object structure. The boundary semantic is then integrated with the initial saliency map generated by dense aggregation strategy from high-level features to guide the saliency prediction in the decoder network. 
Extensive experiments and ablation studies demonstrate the effectiveness of the proposed method and superior performance to other state-of-the-art methods using eight evaluation metrics on two public challenging remote sensing SOD datasets. 
We find that the boundary/edge cue is very beneficial for weakly-supervised salient object detection. To our best knowledge, this is the first work to learn the saliency of remote sensing images from sparse and weak annotations. We hope our research offers the community an opportunity to explore more in this new field.


\bibliographystyle{elsarticle-num} 
\bibliography{main-ISPRS.bib}

\begin{thebibliography}{10}
\expandafter\ifx\csname url\endcsname\relax
  \def\url#1{\texttt{#1}}\fi
\expandafter\ifx\csname urlprefix\endcsname\relax\def\urlprefix{URL }\fi
\expandafter\ifx\csname href\endcsname\relax
  \def\href#1#2{#2} \def\path#1{#1}\fi

\bibitem{borji2019salient}
A.~Borji, M.-M. Cheng, Q.~Hou, H.~Jiang, J.~Li, Salient object detection: A
  survey, Comput. Vis. Media 5~(2) (2019) 117--150.

\bibitem{zhang2019leveraging}
D.~Zhang, J.~Han, L.~Zhao, D.~Meng, Leveraging prior-knowledge for weakly
  supervised object detection under a collaborative self-paced curriculum
  learning framework, Int. J. Comput. Vis. 127~(4) (2019) 363--380.

\bibitem{zhang2017online}
P.~Zhang, T.~Zhuo, W.~Huang, K.~Chen, M.~Kankanhalli, Online object tracking
  based on cnn with spatial-temporal saliency guided sampling, Neurocomputing
  257 (2017) 115--127.

\bibitem{zhao2013unsupervised}
R.~Zhao, W.~Ouyang, X.~Wang, Unsupervised salience learning for person
  re-identification, in: Proc. CVPR, 2013, pp. 3586--3593.

\bibitem{hoyer2019grid}
L.~Hoyer, M.~Munoz, P.~Katiyar, A.~Khoreva, V.~Fischer, Grid saliency for
  context explanations of semantic segmentation, arXiv:1907.13054 (2019).

\bibitem{ramanishka2017top}
V.~Ramanishka, A.~Das, J.~Zhang, K.~Saenko, Top-down visual saliency guided by
  captions, in: Proc. CVPR, 2017, pp. 7206--7215.

\bibitem{hadizadeh2013saliency}
H.~Hadizadeh, I.~V. Baji{\'c}, Saliency-aware video compression, IEEE Trans.
  Image Process. 23~(1) (2013) 19--33.

\bibitem{fan2020camouflaged}
D.-P. Fan, G.-P. Ji, G.~Sun, M.-M. Cheng, J.~Shen, L.~Shao, Camouflaged object
  detection, in: Proc. CVPR, 2020, pp. 2777--2787.

\bibitem{hou2019deeply}
Q.~Hou, M.~Cheng, X.~Hu, A.~Borji, Z.~Tu, P.~Torr, Deeply supervised salient
  object detection with short connections., IEEE Trans. Pattern Anal. Mach.
  Intell. 41~(4) (2019) 815.

\bibitem{zhao2019egnet}
J.-X. Zhao, J.-J. Liu, D.-P. Fan, Y.~Cao, J.~Yang, M.-M. Cheng, Egnet: Edge
  guidance network for salient object detection, in: Proc. ICCV, 2019, pp.
  8779--8788.

\bibitem{pang2020multi}
Y.~Pang, X.~Zhao, L.~Zhang, H.~Lu, Multi-scale interactive network for salient
  object detection, in: Proc. CVPR, 2020, pp. 9413--9422.

\bibitem{zhao2020suppress}
X.~Zhao, Y.~Pang, L.~Zhang, H.~Lu, L.~Zhang, Suppress and balance: A simple
  gated network for salient object detection, in: Proc. ECCV, Springer, 2020,
  pp. 35--51.

\bibitem{zhang2020dense}
Q.~Zhang, R.~Cong, C.~Li, M.-M. Cheng, Y.~Fang, X.~Cao, Y.~Zhao, S.~Kwong,
  Dense attention fluid network for salient object detection in optical remote
  sensing images, IEEE Trans. Image Process. 30 (2020) 1305--1317.

\bibitem{huang2021semantic}
Z.~Huang, H.~Chen, B.~Liu, Z.~Wang, Semantic-guided attention refinement
  network for salient object detection in optical remote sensing images, Remote
  Sens. 13~(11) (2021) 2163.

\bibitem{bearman2016s}
A.~Bearman, O.~Russakovsky, V.~Ferrari, L.~Fei-Fei, What’s the point:
  Semantic segmentation with point supervision, in: ECCV, Springer, 2016, pp.
  549--565.

\bibitem{yu2021structure}
S.~Yu, B.~Zhang, J.~Xiao, E.~G. Lim, Structure-consistent weakly supervised
  salient object detection with local saliency coherence, in: Proc. AAAI, 2021.

\bibitem{zhang2020weakly}
J.~Zhang, X.~Yu, A.~Li, P.~Song, B.~Liu, Y.~Dai, Weakly-supervised salient
  object detection via scribble annotations, in: Proc. CVPR, 2020, pp.
  12546--12555.

\bibitem{li2019nested}
C.~Li, R.~Cong, J.~Hou, S.~Zhang, Y.~Qian, S.~Kwong, Nested network with
  two-stream pyramid for salient object detection in optical remote sensing
  images, IEEE Trans. Geosci. Remote Sens. 57~(11) (2019) 9156--9166.

\bibitem{feng2018novel}
W.~Feng, H.~Sui, J.~Tu, W.~Huang, K.~Sun, A novel change detection approach
  based on visual saliency and random forest from multi-temporal
  high-resolution remote-sensing images, Int. J. Remote Sens. 39~(22) (2018)
  7998--8021.

\bibitem{zhang2017new}
L.~Zhang, J.~Zhang, A new saliency-driven fusion method based on complex
  wavelet transform for remote sensing images, IEEE Geosci. Remote Sens. Lett.
  14~(12) (2017) 2433--2437.

\bibitem{dong2018ship}
C.~Dong, J.~Liu, F.~Xu, Ship detection in optical remote sensing images based
  on saliency and a rotation-invariant descriptor, Remote Sens. 10~(3) (2018)
  400.

\bibitem{zhang2018saliency}
L.~Zhang, Q.~Sun, Saliency detection and region of interest extraction based on
  multi-image common saliency analysis in satellite images, Neurocomputing 283
  (2018) 150--165.

\bibitem{xiang2019mini}
T.-Z. Xiang, G.-S. Xia, L.~Zhang, Mini-unmanned aerial vehicle-based remote
  sensing: techniques, applications, and prospects, IEEE Geoscience and Remote
  Sensing Magazine 7~(3) (2019) 29--63.

\bibitem{zhu2017deep}
X.~X. Zhu, D.~Tuia, L.~Mou, G.-S. Xia, L.~Zhang, F.~Xu, F.~Fraundorfer, Deep
  learning in remote sensing: A comprehensive review and list of resources,
  IEEE Geoscience and Remote Sensing Magazine 5~(4) (2017) 8--36.

\bibitem{wang2017learning}
L.~Wang, H.~Lu, Y.~Wang, M.~Feng, D.~Wang, B.~Yin, X.~Ruan, Learning to detect
  salient objects with image-level supervision, in: Proc. CVPR, 2017, pp.
  136--145.

\bibitem{li2018weakly}
G.~Li, Y.~Xie, L.~Lin, Weakly supervised salient object detection using image
  labels, in: Proc. AAAI, 2018.

\bibitem{liu2021weakly}
Y.~Liu, P.~Wang, Y.~Cao, Z.~Liang, R.~W. Lau, Weakly-supervised salient object
  detection with saliency bounding boxes, IEEE Trans. Image Process. 30 (2021)
  4423--4435.

\bibitem{wang2019boundary}
B.~Wang, G.~Qi, S.~Tang, T.~Zhang, Y.~Wei, L.~Li, Y.~Zhang, Boundary perception
  guidance: a scribble-supervised semantic segmentation approach, in: IJCAI,
  2019.

\bibitem{zhao2021weakly}
W.~Zhao, J.~Zhang, L.~Li, N.~Barnes, N.~Liu, J.~Han, Weakly supervised video
  salient object detection, in: Proc. CVPR, 2021, pp. 16826--16835.

\bibitem{valvano2021learning}
G.~Valvano, A.~Leo, S.~A. Tsaftaris, Learning to segment from scribbles using
  multi-scale adversarial attention gates, IEEE Trans. Med. Imaging (2021).

\bibitem{hua2021semantic}
Y.~Hua, D.~Marcos, L.~Mou, X.~X. Zhu, D.~Tuia, Semantic segmentation of remote
  sensing images with sparse annotations, IEEE Geosci. Remote Sens. Lett.
  (2021).

\bibitem{wei2021scribble}
Y.~Wei, S.~Ji, Scribble-based weakly supervised deep learning for road surface
  extraction from remote sensing images, IEEE Trans. Geosci. Remote Sens.
  (2021).

\bibitem{feng2019attentive}
M.~Feng, H.~Lu, E.~Ding, Attentive feedback network for boundary-aware salient
  object detection, in: Proc. CVPR, 2019, pp. 1623--1632.

\bibitem{zhou2016learning}
B.~Zhou, A.~Khosla, A.~Lapedriza, A.~Oliva, A.~Torralba, Learning deep features
  for discriminative localization, in: Proc. CVPR, 2016, pp. 2921--2929.

\bibitem{wang2021salient}
W.~Wang, Q.~Lai, H.~Fu, J.~Shen, H.~Ling, R.~Yang, Salient object detection in
  the deep learning era: An in-depth survey, IEEE Transactions on Pattern
  Analysis and Machine Intelligence (2021).

\bibitem{fan2021salient}
D.-P. Fan, J.~Zhang, G.~Xu, M.-M. Cheng, L.~Shao, Salient objects in clutter,
  arXiv preprint arXiv:2105.03053 (2021).

\bibitem{deng2018r3net}
Z.~Deng, X.~Hu, L.~Zhu, X.~Xu, J.~Qin, G.~Han, P.-A. Heng, R3net: Recurrent
  residual refinement network for saliency detection, in: IJCAI, AAAI Press,
  2018, pp. 684--690.

\bibitem{hu2018recurrently}
X.~Hu, L.~Zhu, J.~Qin, C.-W. Fu, P.-A. Heng, Recurrently aggregating deep
  features for salient object detection, in: Proc. AAAI, Vol.~32, 2018.

\bibitem{zhao2019pyramid}
T.~Zhao, X.~Wu, Pyramid feature attention network for saliency detection, in:
  Proc. CVPR, 2019, pp. 3085--3094.

\bibitem{zhou2021rgb}
T.~Zhou, D.-P. Fan, M.-M. Cheng, J.~Shen, L.~Shao, Rgb-d salient object
  detection: A survey, Comput. Vis. Media (2021) 1--33.

\bibitem{fan2020rethinking}
D.-P. Fan, Z.~Lin, Z.~Zhang, M.~Zhu, M.-M. Cheng, Rethinking rgb-d salient
  object detection: Models, data sets, and large-scale benchmarks, IEEE Trans.
  Neural Netw. Learn. Syst. 32~(5) (2020) 2075--2089.

\bibitem{huang2021multi}
Z.~Huang, H.-X. Chen, T.~Zhou, Y.-Z. Yang, B.-Y. Liu, Multi-level cross-modal
  interaction network for rgb-d salient object detection, Neurocomputing 452
  (2021) 200--211.

\bibitem{tu2020rgbt}
Z.~Tu, Y.~Ma, Z.~Li, C.~Li, J.~Xu, Y.~Liu, Rgbt salient object detection: A
  large-scale dataset and benchmark, arXiv:2007.03262 (2020).

\bibitem{zhou2021ecffnet}
W.~Zhou, Q.~Guo, J.~Lei, L.~Yu, J.-N. Hwang, Ecffnet: effective and consistent
  feature fusion network for rgb-t salient object detection, IEEE Trans.
  Circuits. Syst. Video Technol. (2021).

\bibitem{zeng2019multi}
Y.~Zeng, Y.~Zhuge, H.~Lu, L.~Zhang, M.~Qian, Y.~Yu, Multi-source weak
  supervision for saliency detection, in: Proc. CVPR, 2019, pp. 6074--6083.

\bibitem{lin2016scribblesup}
D.~Lin, J.~Dai, J.~Jia, K.~He, J.~Sun, Scribblesup: Scribble-supervised
  convolutional networks for semantic segmentation, in: Proc. CVPR, 2016, pp.
  3159--3167.

\bibitem{dorent2020scribble}
R.~Dorent, S.~Joutard, J.~Shapey, S.~Bisdas, N.~Kitchen, R.~Bradford, S.~Saeed,
  M.~Modat, S.~Ourselin, T.~Vercauteren, Scribble-based domain adaptation via
  co-segmentation, in: MICCAI, Springer, 2020, pp. 479--489.

\bibitem{huang2021contrast}
Z.~Huang, H.-X. Chen, T.~Zhou, Y.-Z. Yang, C.-Y. Wang, B.-Y. Liu,
  Contrast-weighted dictionary learning based saliency detection for vhr
  optical remote sensing images, Pattern Recognit. 113 (2021) 107757.

\bibitem{fu2019dual}
J.~Fu, J.~Liu, H.~Tian, Y.~Li, Y.~Bao, Z.~Fang, H.~Lu, Dual attention network
  for scene segmentation, in: Proc. CVPR, 2019, pp. 3146--3154.

\bibitem{tian2020weakly}
X.~Tian, K.~Xu, X.~Yang, B.~Yin, R.~W. Lau, Weakly-supervised salient instance
  detection, arXiv:2009.13898 (2020).

\bibitem{canny1986computational}
J.~Canny, A computational approach to edge detection, IEEE Trans. Pattern Anal.
  Mach. Intell.~(6) (1986) 679--698.

\bibitem{xie2015holistically}
S.~Xie, Z.~Tu, Holistically-nested edge detection, in: Proc. ICCV, 2015, pp.
  1395--1403.

\bibitem{chen2020weakly}
L.~Chen, W.~Wu, C.~Fu, X.~Han, Y.~Zhang, Weakly supervised semantic
  segmentation with boundary exploration, in: ECCV, Springer, 2020, pp.
  347--362.

\bibitem{ronneberger2015u}
O.~Ronneberger, P.~Fischer, T.~Brox, U-net: Convolutional networks for
  biomedical image segmentation, in: MICCAI, Springer, 2015, pp. 234--241.

\bibitem{perazzi2012saliency}
F.~Perazzi, P.~Kr{\"a}henb{\"u}hl, Y.~Pritch, A.~Hornung, Saliency filters:
  Contrast based filtering for salient region detection, in: CVPR, IEEE, 2012,
  pp. 733--740.

\bibitem{achanta2009frequency}
R.~Achanta, S.~Hemami, F.~Estrada, S.~Susstrunk, Frequency-tuned salient region
  detection, in: CVPR, IEEE, 2009, pp. 1597--1604.

\bibitem{fan2018enhanced}
D.-P. Fan, C.~Gong, Y.~Cao, B.~Ren, M.-M. Cheng, A.~Borji, Enhanced-alignment
  measure for binary foreground map evaluation, in: IJCAI, 2018, pp. 698--704.

\bibitem{fan2017structure}
D.-P. Fan, M.-M. Cheng, Y.~Liu, T.~Li, A.~Borji, Structure-measure: A new way
  to evaluate foreground maps, in: ICCV, 2017, pp. 4548--4557.

\bibitem{paszke2017automatic}
A.~Paszke, S.~Gross, S.~Chintala, G.~Chanan, E.~Yang, Z.~DeVito, Z.~Lin,
  A.~Desmaison, L.~Antiga, A.~Lerer, Automatic differentiation in pytorch
  (2017).

\bibitem{he2016deep}
K.~He, X.~Zhang, S.~Ren, J.~Sun, Deep residual learning for image recognition,
  in: Proc. CVPR, 2016, pp. 770--778.

\bibitem{simonyan2014very}
K.~Simonyan, A.~Zisserman, Very deep convolutional networks for large-scale
  image recognition, arXiv:1409.1556 (2014).

\bibitem{liu2019simple}
J.-J. Liu, Q.~Hou, M.-M. Cheng, J.~Feng, J.~Jiang, A simple pooling-based
  design for real-time salient object detection, in: Proc. CVPR, 2019, pp.
  3917--3926.

\bibitem{zhang2017supervision}
D.~Zhang, J.~Han, Y.~Zhang, Supervision by fusion: Towards unsupervised
  learning of deep salient object detector, in: Proc. ICCV, 2017, pp.
  4048--4056.

\bibitem{long2015fully}
J.~Long, E.~Shelhamer, T.~Darrell, Fully convolutional networks for semantic
  segmentation, in: Proc. CVPR, 2015, pp. 3431--3440.

\end{thebibliography}
\end{document}